\documentclass[a4paper]{article}

\usepackage{INTERSPEECH_v2}
\usepackage{xcolor}
\usepackage{multirow}
\usepackage{enumitem}
\usepackage[round-mode=places, round-precision=2]{siunitx}
\usepackage{url}						%Supports url commands

\usepackage[spanish,english]{babel}		%For languages characters and hyphenation
\usepackage[utf8]{inputenc}

\title{\uppercase{Assessing the tolerance of Neural Machine Translation systems against Speech Recognition Errors}}
\name{Nicholas Ruiz$^{1,2}$, Mattia Antonino Di Gangi$^1$, Nicola Bertoldi$^1$, Marcello Federico$^1$}
\address{
  $^1$Fondazione Bruno Kessler, Italy\\
  $^2$Interactions, LLC, USA}
\email{nruiz@interactions.com, \{digangi,bertoldi,federico\}@fbk.eu}

\begin{document}

\maketitle
\begin{abstract}
Machine translation systems are conventionally trained on textual resources that do not model phenomena that occur in spoken language. While the evaluation of neural machine translation systems on textual inputs is actively researched in the literature, little has been discovered about the complexities of translating spoken language data with neural models.
We introduce and motivate interesting problems one faces when considering the translation of automatic speech recognition (ASR) outputs on neural machine translation (NMT) systems.
We test the robustness of sentence encoding approaches for NMT encoder-decoder modeling, focusing on word-based over byte-pair encoding.
% Treating the outputs of multiple ASR systems as a source of noise that impacts translation quality,
We compare the translation of utterances containing ASR errors in  state-of-the-art NMT encoder-decoder systems against a strong phrase-based machine translation baseline in order to better understand which phenomena present in ASR outputs are better represented under the NMT framework than approaches that represent translation as a linear model.


\end{abstract}

%
% Author: Nicholas Ruiz
% Description: Neural Spoken Language Translation Adaptation
%
\section{Introduction}

A substantial amount of progress has been made in Neural Machine Translation (NMT) for text documents. Research has shown that the encoder-decoder model with an attention mechanism generates high quality translations that exploit long range dependencies in an input sentence \cite{Bahdanau:15}. 
While NMT has proven to yield significant improvements for text translation over log-linear approaches to MT such as phrase-based machine translation (PBMT), it has yet to be shown the extent to which gains purported in the literature generalize to the scenario of spoken language translation (SLT), where the input sequence may be corrupted by noise in the audio signal and uncertainties during automatic speech recognition (ASR) decoding. 
Are NMT models implicitly better at modeling and mitigating ASR errors than the former state-of-the-art approaches to machine translation?
As a preliminary work, we analyze the impact of ASR errors on neural machine translation quality by studying the properties of the translations provided by an encoder-decoder NMT system with an attention mechanism, against a strong baseline PBMT system that rivals the translation quality of Google Translate\texttrademark~ on TED talks.

We address the following questions regarding NMT:

\begin{enumerate}
\item How do NMT systems react when ASR transcripts are provided as input?
\item Do ASR error types in word error alignments impact SLT quality the same for NMT as PBMT? Or is NMT implicitly more tolerant against ASR errors?
\item Which types of sentences does NMT handle better than PBMT, and vice-versa?
% \item What causes NMT and PBMT systems to generate the same translation hypothesis, in the presence of speech recognition errors?
% \item Does BPE translate sentences with ASR errors corresponding to low frequency words better than full-word systems?
\end{enumerate}

% Ruiz et al., 2014 and Tsvetkov et al, 2013 explored noisy channel modeling techniques to adapt phrase-based MT systems by generating synthetic data. In this work, we will explore similar techniques to increase the amount of speech-like training data by synthesizing ASR errors on text data. 
To address these questions, we explore the impact of feeding ASR hypotheses, which may contain noise, disfluencies, and different representations on the surface text, to a NMT system that has been trained on TED talk transcripts that do not reflect the noisy conditions of ASR. Our experimental framework is similar to that of \cite{Ruiz:14:AMTA,Ruiz:15:ASRU}, with the addition of a ranking experiment to evaluate the quality of NMT against our PBMT baseline.
These experiments are intended as an initial analysis with the purpose to suggesting directions to focus on in the future.

\section{Neural versus Statistical MT}
\label{sec:nmt-related}
Before beginning our analysis, we summarize some of the biggest differences between NMT and other forms of statistical machine translation, such as PBMT.

\cite{Bentivogli:16:nmtCompare} compare neural machine translation against three top- performing statistical machine translation systems in the TED talk machine translation track from IWSLT 2015.\footnote{The International Workshop of Spoken Language Translation.}
The evaluation set consists of 600 sentences and 10,000 words, post-edited by five professional translators.
In addition to reporting a 26\% relative improvement in multi-reference TER (mTER), \cite{Luong:15:IWSLT}'s encoder-decoder attention-based NMT system trained on full words outperformed state of the art statistical machine translation (SMT) systems on English-German, a language pair known to have issues with morphology and whose syntax differs significantly from English in subordinate clauses.
% Although the NMT system outperformed every PBMT system, regardless of the sentence length, \citeauthor{Bentivogli:16:nmtCompare} confirmed \cite{Cho:14b}'s findings that translation quality deteriorates rapidly as the sentence length approaches 35 words.
\cite{Bentivogli:16:nmtCompare}'s analysis yields the following observations:
\begin{itemize}
\item \textit{Precision versus Sentence length}: Although NMT outperformed every comparable log-linear MT system, they confirmed \cite{Cho:14b}'s findings that translation quality deteriorates rapidly as the sentence length approaches 35 words.
\item \textit{Morphology}: NMT translations have better case, gender and number agreement than PBMT systems.
\item \textit{Lexical choice}: NMT made 17\% fewer lexical errors than any PBMT system.
\item \textit{Word order}: NMT yielded fewer \textit{shift} errors in TER alignments than any SMT system. NMT yielded significantly higher Kendall Reordering Score (KRS) \cite{Birch:09} values than any PBMT system. NMT generated 70\% fewer verb order errors than the next-best hybrid phrase and syntax-based system.
\end{itemize}

Several SMT modeling challenges are exacerbated in NMT.
While log-linear SMT translation models can handle large word vocabularies, NMT systems require careful modeling to balance vocabulary coverage and network size, since each token introduced increases the size of its hidden layers.
% A number of researchers outline the translation of out-of-vocabulary words as the words most under-performing objective.
Because of this constraint,
\cite{Hirschmann:16:nmtHard} observe that only 69\% of German nouns are covered with 30,000 words on English-German WMT 2014 system.
\footnote{WMT 2014 training data consists primarily of news texts, European parliament proceedings, and web crawled data. \url{http://www.statmt.org/wmt14/translation-task.html}}
Although noun compound splitting works well for German$\rightarrow$English, English$\rightarrow$German model performance not improve significantly.
In particular named entities (e.g. persons, organizations, and locations) are underrepresented.

On the other hand, NMT has the ability to model subword units such as characters \cite{Chung:16:character} or coarser grained segmentations of low frequency words \cite{Sennrich:16a} without substantial changes to the system architecture, unlike other SMT approaches.
% A PBMT system such as that described by \cite{Koehn:07a} requires strong constraints on the distance of reordering operations, as well as the number of tokens permissible in the source or target language side of the translation model. Likewise, differences in the orthographic representation of the source and target strings can be problematic, for example, in Japanese-English or Mandarin-English phrase-based MT \cite{Chang:08:chinese,Wang:07:chinese}. 
% Works such as \cite{Nakov:12,Neubig:13:substring} attempt to build SMT translation models that combine full words and substring units.
% The downside of fine-grained segmentations such in as character-based models for NMT is that the length of the input affects training and decoding time, as both the attention and hidden state models are modeled by recurrent neural network architectures modeling long distance dependencies the observations in an input string.
% Nevertheless, the representation of subword units allow the attention model to decide the length and grain of the input sequence that is useful for decoding each target position.

\cite{Firat:16:multi} have additionally demonstrated NMT's ability to translate between multiple language pairs with a neural translation model trained with a single attention mechanism.
% While there have been attempts to model multilingual translation in traditional forms of statistical machine using multi-source machine translation \cite{Och:01:multisource,Schroeder:09:lattice} or pivoting \cite{Cohn:07,Bertoldi:08}, most phrase-based and hierarchical MT systems require training on individual language pairs.

Although NMT models translate with higher precision, models are slow to train even with the most powerful GPUs -- often taking weeks for the strongest systems to complete training. On the other hand, large order PBMT systems trained in the ModernMT framework\footnote{\url{http://www.modernmt.eu}} may be trained within a few hours and can be adapted in near real-time with translation memories containing post-editions by professional translators.

\section{Research Methodology}
% % The goal of this chapter is to understand the effects of ASR errors on NMT and to measure if NMT handles ASR error types in a similar manner as we showed in the PBMT analysis in Chapter~\ref{chap:sltQuality}.
% We train several NMT models using the data from the IWSLT 2013 TED talk speech translation task. We focus on the English-French translation task. In a similar evaluation framework as in Chapter~\ref{chap:sltQuality}, we analyze the effects of applying 8 different ASR system outputs to a number of phrase-based and NMT systems. We look for methods that perform consistently well for the different quality levels provided by the ASR systems.

% We analyze the effects of using different input-output encodings with ASR outputs to determine which representation is more tolerant to noise in the input string. 

% In the typical cascading approach to SLT, the ASR and MT components are trained separately.
% Since there is a paucity of training data available that simultaneously comprises source-language audio recordings, ASR references, and target language translations, engineers will typically train MT systems on text data, which do not match well with the domain and style of speech data.
% In this work we analyze the differences between NMT and PBMT error types for the SLT task. As part of our analysis, we apply the ASR error analysis approach of \cite{Ruiz:14:AMTA,Ruiz:15:ASRU} to better understand the impact of ASR errors on neural translation models.
% These experiments are intended as an intial analysis with the purpose to suggesting directions to focus on in the future.

Similar to our experimental framework in \cite{Ruiz:14:AMTA,Ruiz:15:ASRU}, 
% we perform our experiments on an intersection of the ASR and MT results of the IWSLT 2013 evaluation campaign \cite{Cettolo:2013:IWSLT}, which focused on the translation of TED talks.
we collect English ASR hypotheses from the eight submissions on the tst$_{2012}$ test set in the IWSLT 2013 TED talk ASR track \cite{Cettolo:2013:IWSLT}.
Coupled with reference translations from the MT track, we construct a dataset consisting of the eight English ASR hypotheses for 1,124 utterances, a single unpunctuated reference transcript from the ASR track, and the reference translations from the English-French MT track.
The English ASR hypotheses and reference transcript are normalized and punctuated according to the same approach as described in \cite{Ruiz:15:ASRU}.
We use both BLEU \cite{Papineni:02} and Translation Edit Rate (TER) \cite{Snover:06} as global evaluation metrics. TER and $\Delta$TER over gold standard ASR outputs are used to observe sentence-level trends. We compute automatic translation scores, sentence-level system ranking, and take a closer look at the types of errors observed in the data.
%
% In the sections b
Below, we briefly describe the MT systems used in this experiment.

\subsection{Neural MT system}
Our NMT system is based on FBK's primary MT submission to the IWSLT 2016 evaluation for English-French TED talk translation \cite{Farajian:16:IWSLT}.
The system is based on the sequence-to-sequence encoder-decoder architecture proposed in \cite{Bahdanau:15} and further developed by  \cite{Luong:15:IWSLT, Sennrich:16a}. 
The system is trained on full-word text units to allow a direct comparison with our PBMT counterpart. We refer to this system as \textsc{neural} for the remainder of our experiments.

\subsection{Phrase-based MT system}
\label{sec:ModernMT}
Our phrase-based MT system (which we refer to as \textsc{mmt}) is built upon the ModernMT framework: an extension of the phrase-based MT framework introduced in \cite{Koehn:07a} that enables context-aware translation for fast and incremental training.
Context-aware translation is achieved by the partitioning of the training data into homogeneous domains by a \textit{context analyzer} (CA), which permits the rapid construction and interpolation of domain-specific translation, language, and reordering models based on the context received during a decoding run. It also permits the underlying models to be modified online. 
The decoder also exploits other standard features (phrase, word, distortion, and unknown word penalties) and performs cube-pruning search.
% \NB{
% Fast and incremental training is achieved by means of probabilistic models, which computes their scores on-the-fly, and data structures, which contain the minimal information required by the models, and can be updated when new training datas occur.
% }
A detailed description of the ModernMT project can be found in \cite{Bertoldi:16:ModernMT}.

\section{SLT Evaluation}
\begin{table}[tb]
\centering
\footnotesize
\begin{tabular}{l|r|rr|rr}
\hline
\hline
 &     & \multicolumn{2}{c}{\textsc{mmt}}       &  \multicolumn{2}{c}{\textsc{neural}}       \\
ASR system           &  WER $\downarrow$  & BLEU $\uparrow$ & TER $\downarrow$                         & BLEU $\uparrow$ & TER $\downarrow$   \\
\hline
gold                 &  0.0  & 43.4 & 39.5                       & 47.9 & 35.4 \\
\hline
fbk                  & 16.5  & 35.6 & 48.4                       & 38.5 & 45.6 \\
kit                  & 10.1  & 38.1 & 45.0                       & 41.8 & 41.7 \\
mitll                & 11.4  & 37.7 & 45.8                       & 41.4 & 42.4 \\
naist                & 10.6  & 38.1 & 45.0                       & 41.8 & 41.5 \\
nict                 &  9.2  & 38.7 & 44.7                       & 42.5 & 41.1 \\
prke                 & 16.6  & 34.9 & 48.7                       & 38.1 & 45.8 \\
rwth                 & 11.7  & 37.2 & 46.1                       & 41.4 & 42.3 \\
uedin                & 12.3  & 37.3 & 46.1                       & 40.8 & 42.9 \\
\hline
\end{tabular}
\caption[A comparison of Neural MT versus Phrase-based MT on the SLT evaluation of TED talks (tst$_{2012}$) from the IWSLT 2013 evaluation campaign.]{A comparison of Neural MT versus Phrase-based MT on the SLT evaluation of TED talks (tst$_{2012}$) from the IWSLT 2013 evaluation campaign \cite{Cettolo:2013:IWSLT}. Evaluation results are compared to a \textit{gold} standard that assumes that no ASR errors occur.\vspace{-0.5cm}}
\label{tbl:nmt-eval}
\end{table}

We first report the translation results on the evaluation task in Table \ref{tbl:nmt-eval}. NMT outperforms our best PBMT system by 4.5 BLEU in the absence of ASR errors (\textit{gold}) and by approximately 3 BLEU across all ASR hypothesis inputs.
Overall, the introduction of ASR errors results in
decreases in BLEU by $5.5 (\pm 0.8)$ and $5.4 (\pm 0.8)$
and
TER increases of $6.0 (\pm 0.9)$ and $6.2 (\pm 0.9)$ for \textsc{mmt} and \textsc{neural}, respectively.

Table \ref{nmt-eval-rank-avg} provides the average sentence-level TER and $\Delta$TER scores, which report the degradation of SLT quality by the presence of ASR errors.
Although the average TER scores from the \textsc{mmt} outputs are higher, the $\Delta$TER scores are lower than their \textsc{neural} counterparts, implying that the \textsc{mmt} SLT outputs are closer to their gold standard MT outputs.
This may suggest that NMT is more sensitive to local changes to an input caused by minor ASR errors.

\begin{table}[tb]
\centering
\footnotesize
\begin{tabular}{l|rr|rr|rr}
\hline\hline
      & \multicolumn{2}{c}{\textsc{mmt}} &  \multicolumn{2}{c}{\textsc{neural}}& \multicolumn{2}{c}{\textsc{difference}} \\%&  &  \\
SysID & TER $\downarrow$ & $\Delta$TER $\downarrow$ & TER & $\Delta$TER & TER & $\Delta$TER \\
\hline
gold  & 39.6     & 0.0             & 35.6        & 0.0                & -4.0      & 0.0              \\
fbk   & 49.3     & 9.7             & 46.6        & 11.0               & -2.7      & 1.3              \\
kit   & 45.9     & 6.3             & 42.7        & 7.1                & -3.2      & 0.8              \\
mitll & 46.8     & 7.2             & 43.7        & 8.1                & -3.1      & 0.9              \\
naist & 45.6     & 6.1             & 42.1        & 6.5                & -3.5      & 0.5              \\
nict  & 45.1     & 5.5             & 41.9        & 6.3                & -3.1      & 0.9              \\
prke  & 49.4     & 9.8             & 46.5        & 10.9               & -2.9      & 1.1              \\
rwth  & 47.0     & 7.4             & 43.2        & 7.6                & -3.8      & 0.2              \\
uedin & 46.7     & 7.1             & 43.8        & 8.2                & -2.9      & 1.1      \\
\hline       
\end{tabular}
\caption[Average utterance-level translation TER and $\Delta$TER scores for the MMT and Neural MT systems.]{Average utterance-level translation TER and $\Delta$TER scores for the MMT and Neural MT systems. The average Neural MT TER scores are 3\% better than the PBMT counterpart.\vspace{-0.5cm}}
\label{nmt-eval-rank-avg}
\end{table}

% How many ties are shared in common between the gold ASR translations and each ASR system hypothesis' translations? The intersection between these sets lets us know which utterances are potentially more difficult for NMT or PBMT.

\subsection{MT system ranking}

Are there 
ASR error conditions in which PBMT remains a better solution than NMT, and if so, what are the properties of these utterances that makes them difficult for NMT?
We take a closer look at the sentence-level translation scores by ranking the performance of each MT system on the utterances where ASR errors exist, in order to understand how each MT system handles noisy input. 
% find the segments that \textsc{mmt} translates better, as opposed those that \textsc{neural} handles better.
For each utterance, we rank the systems based on their the sentence-level TER scores computed on their translation outputs over each ASR hypothesis. We also mark ties, in which both systems yield the same TER score. Results containing the counts and percentage of wins by MT system are provided in Table \ref{tbl:nmt-eval-rank}.

\begin{table}[tb]
\centering
\footnotesize
\begin{tabular}{l|lrr|rr}
\hline\hline
                       &       \multicolumn{3}{c|}{SLT Rank by sentence}                    & \multicolumn{2}{c}{TER (avg) $\downarrow$} \\
Lab                  & Winner & Count & Percentage & \textsc{mmt}     & \textsc{neural}     \\
\hline\multirow{3}{*}{fbk}   & \textsc{mmt} & 257   & 32.4       & 51.1     & 64.0     \\
                       & \textsc{neural} & 373   & 47.1       & 58.9     & 44.5     \\
                       & Tie    & 162   & 20.5       & 54.1     & 54.1     \\
\hline\multirow{3}{*}{kit}   & \textsc{mmt} & 213   & 30.6       & 46.3     & 59.1     \\
                       & \textsc{neural} & 347   & 49.9       & 55.0     & 41.1     \\
                       & Tie    & 135   & 19.4       & 52.9     & 52.9     \\
\hline\multirow{3}{*}{mitll} & \textsc{mmt} & 194   & 27.6       & 48.4     & 61.4     \\
                       & \textsc{neural} & 351   & 49.9       & 55.5     & 41.5     \\
                       & Tie    & 159   & 22.6       & 52.2     & 52.2     \\
\hline\multirow{3}{*}{naist} & \textsc{mmt} & 189   & 28.3       & 43.9     & 56.5     \\
                       & \textsc{neural} & 342   & 51.2       & 54.8     & 41.2     \\
                       & Tie    & 137   & 20.5       & 52.5     & 52.5     \\
\hline\multirow{3}{*}{nict}  & \textsc{mmt} & 184   & 31.8       & 46.3     & 58.4     \\
                       & \textsc{neural} & 286   & 49.4       & 54.0     & 40.8     \\
                       & Tie    & 109   & 18.8       & 56.0     & 56.0     \\
\hline\multirow{3}{*}{prke}  & \textsc{mmt} & 256   & 31.6       & 48.0     & 60.3     \\
                       & \textsc{neural} & 378   & 46.7       & 57.7     & 44.1     \\
                       & Tie    & 175   & 21.6       & 55.8     & 55.8     \\
\hline\multirow{3}{*}{rwth}  & \textsc{mmt} & 221   & 29.9       & 47.0     & 59.2     \\
                       & \textsc{neural} & 383   & 51.8       & 55.5     & 41.3     \\
                       & Tie    & 135   & 18.3       & 55.0     & 55.0     \\
\hline\multirow{3}{*}{uedin} & \textsc{mmt} & 219   & 30.6       & 47.9     & 59.2     \\
                       & \textsc{neural} & 348   & 48.7       & 56.5     & 42.8     \\
                       & Tie    & 148   & 20.7       & 52.2     & 52.2 \\
\hline
\end{tabular}

\caption[Ranked evaluation of translated SLT utterances.]{ Ranked evaluation of the SLT utterances containing ASR errors in tst$_{2012}$. 
(Left) Counts of the winner decisions and the percentage of all of the decisions that were influenced by ASR errors.
(Right) Mean TER scores across each sentence in the ranked set.
The remainder of winner decisions are made on error-free ASR transcripts. \vspace{-0.6cm}}
\label{tbl:nmt-eval-rank}
\end{table}

The \textsc{neural} and \textsc{mmt} scores are tied on over 20\% of the utterances.
For the better performing ASR systems (e.g. NICT, KIT), we observe a slightly higher proportion of utterances with better NMT translations and a reduced number of ties.
On the right-hand side of Table~\ref{tbl:nmt-eval-rank} we report the average TER scores within each ranking partition of the data. For example, for the utterances that are translated better by \textsc{mmt}, we observe that the average TER scores for \textsc{neural} have an absolute average improvement of 10\% in TER over \textsc{mmt}. The converse is also true, suggesting that there is a subset of utterances that \textsc{mmt} translates better than \textsc{neural}. 
% We now focus on these utterances and determine whether the ranking decisions made under noisy scenarios matches the scenario where the utterance was recognized perfectly.

% \begin{figure}[tb]
% \centering
% \includegraphics[width=1.0\textwidth]{fig/slt_errors}
% \caption{Changes in MT system rankings as ASR errors are introduced.
% Tuples are labeled by (MT rank, SLT rank).}
% \label{fig:nmt-slt-errors}
% \end{figure}

We look into the translation errors caused by ASR errors by plotting the changes in MT system ranking as we shift from the perfect ASR scenario to the actual ASR results from the evaluation (Table~\ref{tbl:asr-rank}).
Across all ASR outputs, 70.2\% of the MT evaluation ranking decisions remain the same when ASR errors create noisy input.
The \textsc{neural} model retains a higher rank 7.5\% more often than \textsc{mmt} as ASR errors are introduced.
Ranking ties remain 55.5\% of the time.
Of the remaining, the \textsc{neural} model outperforms \textsc{mmt} 5.5\% more often in the presence of ASR errors (25.0\% versus 19.5\%).
These results confirm that at the corpus level NMT produces higher scoring translations in the presence of ASR errors.

% the distribution of wins reported in Table \ref{tbl:nmt-eval-rank} suggests that systems with more ASR errors have an increased percentage of sentences that are either scored higher by \textsc{mmt} or are ties.
% 76.5\% of the sentences with a better \textsc{neural} translations preserve their rankings as ASR errors are introduced, a 7.5\% increase over \textsc{MMT} translations.
% Of the 29.8\% of changes, an average of 40.8\% favor \textsc{neural} and 29.8\% favor \textsc{mmt} across all systems. 
% There are fewer cases of mismatches between the gold ranking decisions as the WER of the ASR system decreases.

\begin{table}[tb]
\footnotesize
\centering
\begin{tabular}{l|l|r|r}
\hline\hline
Gold Winner	&	ASR Winner	&	Absolute \%	&	Relative \%	\\
\hline
\multirow{3}{*}{\textsc{mmt}}	&	\textsc{mmt}	&	19.8	&	69.0	\\
	&	\textsc{neural}	&	4.6	&	16.1	\\
	&	Tie	&	4.3	&	14.9	\\
	\hline
\multirow{3}{*}{\textsc{neural}}	&	\textsc{mmt}	&	6.4	&	12.4	\\
	&	\textsc{neural}	&	39.4	&	76.5	\\
	&	Tie	&	5.7	&	11.1	\\
	\hline
\multirow{3}{*}{Tie}	&	\textsc{mmt}	&	3.9	&	19.5	\\
	&	\textsc{neural}	&	5.0	&	25.0	\\
	&	Tie	&	11.0	&	55.5	\\
\hline
\end{tabular}
\caption{Changes in sentence-level TER rankings as ASR errors are introduced.\vspace{-0.5cm}}
\label{tbl:asr-rank}
\end{table}
% \textsc{neural} produces better scoring translations on over 47\% of the utterances.

\subsection{Translation examples}
Although NMT may outperform phrase-based SMT, our experiment shows that \textsc{mmt} still outperforms \textsc{neural} 30.1\% of the time.
In order to understand this behavior, we provide three examples of key differences between in how \textsc{neural} and \textsc{mmt} mitigate FBK's ASR errors (Fig.~\ref{fig:nmt-examples}).
In utterance U4, \textsc{neural} is missing the translation of two content words from its vocabulary. In the absence of errors \textsc{neural}$_\textit{gold}$ passes the source word ``embody'' through to its output without translating it. During ASR, ``embody'' is misrecognized as ``body'', which is also passed through without a translation. 
We find it strange that ``body'' was not translated as ``corps'', given that other utterances containing ``body'' receive that translation.
After investigating further, we came across other cases of gold transcripts where ``body'' was not translated at all.
Utterances U212, U214, and U242 have the phrase ``body architect'', but only U212 has a translation for the word ``body'':
\begin{itemize}
\item[] I call myself a body architect.\\ \textit{je m'appelle un corps architecte.}
\item[] As a body architect, I'm fascinated with the human body \\ \textit{en tant qu'architecte, je me suis retrouvé avec le corps humain }
\item[] As a body architect, I've created \\ \textit{en tant qu'architecte , j'ai créé}
\end{itemize}
It is likely that \textit{NMT may not be able to translate contextual patterns it hasn't observed before.}
\textsc{mmt} on the other hand provides valid translation for both words; although the meaning of the sentence is lost due to the translation of ASR errors. A PBMT system will translate phrases consistently, as long as there is not another overlapping phrase pair in the translation model that leads to a path in the search graph with a higher score.

Utterance U85 in the TED talk test set shows longer range effects of ASR errors on translation in NMT. FBK's ASR recognized the utterance as ``But when I \textbf{step} back, I felt myself at the cold, hard center of a perfect storm.'' In the translation of ASR$_\textit{gold}$, both MT systems translate the expression ``stepped back'' in the sense of ``returned''. \textsc{mmt}$_\textit{gold}$ reorders ``centre'' incorrectly.
ASR$_\textit{hyp}$ has a single error where the past tense suffix ``-ed'' on ``step'' was lost.
\textsc{neural}$_\textit{ASR}$ provides an adequate translation as ``je recule'', but in the process, the attention mechanism seems to have taken the incorrect source word and translation as context that corrupts the remainder of the translation.
While \textsc{mmt}$_\textit{ASR}$ makes a translation error at the beginning of the sense, the remainder of the translated sentence remains the same as its gold translation. 
This suggests that \textit{ASR errors may have longer range effects on NMT systems} in languages that are even observable in sentences that lack long distance dependencies.

Utterance U296 demonstrates an example where \textit{misrecognitions of short function words can cause the duplication of content words in NMT}. While \textsc{mmt} handles the misrecognition ``and''$\Rightarrow$``an'' by backing off by translating it independently from other phrases in the sentence,
\textsc{neural}, attaches ``photo'' both to the article ``an'' and additionally outputs ``photo'' at its usual position.
As innocuous closed-class word errors that occur often in ASR, this could yield a significant problem in NMT.

% Utterance U907 demonstrates how \textit{NMT is capable of modeling articles that are missing from speech recognition outputs}. 
% \begin{itemize}
% \item ``à l' aise'': at ease vs comfortable
% \item NMT modeling missing articles.
% \end{itemize}

\begin{figure}[tb]
\centering
\resizebox{\linewidth}{!}{%
\begin{tabular}{l|p{5cm}|rr}
\hline\hline
&&TER & $\Delta$TER \\

\hline
ASR$_{gold}$                     & I embody the central paradox.   &  & (U4) \\
ASR$_{hyp}$                      & I \textbf{body} the central paradox.   & & \\
Translation                              & j' incarne le paradoxe central .   &  \\
\hline
\textsc{mmt}$_\textrm{ASR}$             & je \textbf{corps} \textbf{au} paradoxe central .   &  50.0 & 50.0 \\
\textsc{neural}$_\textrm{ASR}$        & je \textbf{body} le paradoxe central .   &  33.33 & 16.66\\
\hline
\textsc{mmt}$_{gold}$            & j' incarne le paradoxe central .   &  0.0 &  \\
\textsc{neural}$_{gold}$       & j' \textbf{embody} le paradoxe central .   &  16.67 & \\
\hline

%\hline
%ASR$_{gold}$                     & In those heady days, we were experimenting with chat rooms and online virtual communities.   &  & (U9) \\
%ASR$_{hyp}$                      & In those heady days, we were experimenting with chat rooms and online virtual communities.   & & \\
%\hline
%Trans                              & en ces jours grisants , nous expérimentions avec salles de chat en ligne et des communautés virtuelles .   &  \\
%\textsc{mmt}$_\textrm{ASR}$             & en ces jours grisants , nous expérimentions avec des salles de chat et communautés virtuelles en ligne .   &  11.11 & 0.0 \\
%\textsc{neural}$_\textrm{ASR}$        & dans ces journées heady , nous faisions des expériences avec des salles de chat et des communautés virtuelles en ligne .   &  44.44 & 0.0\\
%\hline
%\textsc{mmt}$_{gold}$            & en ces jours grisants , nous expérimentions avec des salles de chat et communautés virtuelles en ligne .   &  11.11 &  \\
%\textsc{neural}$_{gold}$       & dans ces journées heady , nous faisions des expériences avec des salles de chat et des communautés virtuelles en ligne .   &  44.44 & \\

\hline

%\hline
%ASR$_{gold}$                     & Getting that text was like getting a hug.   &  & (U3) \\
%ASR$_{hyp}$                      & Getting that texts it was like getting a hug and so there you have it.   & & \\
%\hline
%Trans                              & recevoir ce texto c' était comme recevoir un câlin .   &  \\
%\textsc{mmt}$_\textrm{ASR}$             & pour que les textes c' était comme avoir un câlin et alors là vous l' avez .   &  110.0 & 70.0 \\
%\textsc{neural}$_\textrm{ASR}$        & avec ces textes , c' était comme prendre un câlin et donc vous l' avez .   &  100.0 & 60.0\\
%\hline
%\textsc{mmt}$_{gold}$            & avoir ce texte était comme avoir un câlin .   &  40.0 &  \\
%\textsc{neural}$_{gold}$       & obtenir ce texte était comme prendre un câlin .   &  40.0 & \\
%\hline

\hline
ASR$_{gold}$                     & But when I stepped back, I felt myself at the cold, hard center of a perfect storm.   &  & (U85) \\
ASR$_{hyp}$                      & But when I \textbf{step} back, I felt myself at the cold, hard center of a perfect storm.   & & \\
Translation                              & mais quand j' ai pris du recul , je me suis sentie au centre froid , et dur d' une tempête parfaite .   &  \\
\hline
% \textsc{mmt}$_\textrm{ASR}$             & mais quand j' ai \textit{[]} du recul , je me sentais au froid , dur centre d' une tempête parfaite .   &  21.74 & -17.39 \\
\textsc{neural}$_\textrm{ASR}$        & mais quand \textbf{je recule} , je me sentais \textbf{dans le froid et le centre} d' une tempête parfaite .   &  47.83 & 13.05\\
% \hline
% \textsc{mmt}$_{gold}$            & mais quand \textbf{je suis revenu} , je me sentais \textbf{au froid} , dur centre d' une tempête parfaite .   &  39.13 &  \\
\textsc{neural}$_{gold}$       & mais quand \textbf{je suis revenu} , je me sentais au centre froid et dur d' une tempête parfaite .   &  34.78 & \\
\hline

\hline
ASR$_{gold}$                     & And he emailed me this picture.   &  & (U296) \\
ASR$_{hyp}$                      & An emailed me this picture.   & & \\
\hline
Translation                              & il m' a envoyé cette photo .   &  \\
\textsc{mmt}$_\textrm{ASR}$             & \textbf{un} m' a envoyé cette photo .   &  14.29 & 0.0 \\
\textsc{neural}$_\textrm{ASR}$        & \textbf{une photo} m' a envoyé cette photo .   &  28.57 & 14.28\\
\hline
\textsc{mmt}$_{gold}$            & et il m' a envoyé cette photo .   &  14.29 &  \\
\textsc{neural}$_{gold}$       & et il m' a envoyé cette photo .   &  14.29 & \\
\hline

\end{tabular}%
}
\caption[Three examples of changes in NMT errors caused by ASR errors.]{Three examples of changes in NMT errors (\textsc{neural}) caused by ASR errors:
(1) the effects of unobserved context;
(2) long distance effects of local ASR errors;
(3) duplication of content words caused by substitution errors on short function words.
Alternative translations are provided by \textsc{mmt}.
% TER and $\Delta$TER scores are reported for each sentence translated by \textsc{neural} and \textsc{mmt}.
% U907 demonstrates NMT's ability to model articles missing from ASR output.
\vspace{-0.3cm}%
}
\label{fig:nmt-examples}
\end{figure}

\section{Mixed-effects analysis and error distribution}
In order to quantify the effects of ASR errors on each system, 
% Here we provide some initial observations on the differences in how \textsc{neural} and \textsc{mmt} handle ASR errors. 
% We 
we build linear mixed-effects models \cite{Searle:1974:mixed} in a similar manner to our mixed-effects analysis in \cite{Ruiz:14:AMTA,Ruiz:15:ASRU}.
We construct two sets of mixed-effects models, using the word error rate scores of the 8 ASR hypotheses as independent variables and the resulting increase in translation errors $\Delta$TER as the response variable. The models contain random effect intercepts that account for the variance associated with the ASR system (SysID), the intrinsic difficulty of translating a given utterance (UttID), and a random effects slope accounting for the variability of word error rate scores (WER) across systems.
Instead of treating each different MT system as a random effect in a joint mixed-effect model, 
we construct a mixed-effects model for each MT system with the purpose of comparing the degree to which each ASR error type explains the increase in translation difficulty.
The models are built using R \cite{R:2013} and the \textit{lme4} library \cite{R:lme4}.
The fixed-effects coefficients and the variance of the random effects for each model are shown in Table \ref{tbl:nmtAnalysis-mixed}.

Our first models (\textit{WER-only}) focus on the effects of the global WER score on translation quality ($\Delta$TER). Our fitted models claim that each point of WER yields approximately the same change in $\Delta$TER for \textsc{neural} ($0.61 \pm 0.020$) and \textsc{mmt} ($0.56 \pm 0.019$).

Our second models (WER$_\textit{basic}$) break WER into its \textbf{Sub}stitution, \textbf{Del}etion, and \textbf{Ins}ertion error alignments, each being normalized by the length of the reference transcript. 
% Although we propose the use of phonetic substitution spans in our analysis of \cite{Ruiz:15:ASRU}, we leave them out here because these preliminary experiments do not account for the syntactic properties of the errors.\footnote{The phonetically-oriented alignment approach of \cite{Ruiz:15:ASRU} is useful for future analyses that focus the linguistic properties of the error types, but requires more precise word-level alignments than offered by the conventional bag of alignment errors computed in conventional WER. 
% We reserve the analysis of aligned errors for future exploration. }
According to the fixed effects of the model, insertion errors have a greater impact on translation quality in NMT than deletions.
More importantly, substitution errors have a significantly stronger impact in NMT on translation quality,
which reflects the behavior we observe in the translation examples from Fig.~\ref{fig:nmt-examples}.
MMT appears to be affected by insertion and deletion error types equally.

\begin{table}[tb]
\centering
% \footnotesize
\resizebox{\linewidth}{!}{%
\begin{tabular}{l|rr|rr} %|lrr|rrr}
\hline\hline
              \multicolumn{5}{c}{WER-only (null model)}       \\   
\multicolumn{1}{c}{}     & \multicolumn{2}{c}{\textsc{neural}}           & \multicolumn{2}{c}{\textsc{mmt}}       \\ 
Fixed effects  & $\beta$	& Std. Error            & $\beta$    & Std. Error \\ \hline
(Intercept)       & \num{4.35e-03} & \num{2.68e-03} & \num{-2.08e-05} & \num{1.92e-03} \\ 
WER               & \num{6.09e-01}  & \num{1.98e-02} $\bullet$& \num{5.58e-01}  & \num{1.85e-02} $\bullet$\\ 
\hline
Random effects & Variance   & Std. Dev.          & Variance  & Std. Dev. \\
\hline
UttID (Intercept) & \num{0.005636} & \num{0.07507}  & \num{0.002563}  & \num{0.05063}  \\ 
WER               & \num{0.232551} & \num{0.48224}  & \num{0.223657}  & \num{0.47292}  \\ 
SysID (Intercept) & 0        & 0        & 0         & 0        \\ 
Residual          & \num{0.005451} & \num{0.07383}  & \num{0.003823}  & \num{0.06183}  \\ 
 \hline
 \multicolumn{5}{c}{} \\
 \multicolumn{5}{c}{WER$_\textit{basic}$ (Levenshtein alignment errors)}           \\
\multicolumn{1}{c}{}            & \multicolumn{2}{c}{\textsc{neural}}          &           \multicolumn{2}{c}{\textsc{mmt}}                \\
Fixed effects & $\beta$    & Std. Error            & $\beta$    & Std. Error           \\ \hline
 (Intercept) & \num{4.87e-03} & \num{2.69e-03} &            \num{-5.76e-05} & \num{1.93e-03} \\
 Sub & \num{6.80e-01} & \num{2.10e-02} $\bullet$&            \num{5.35e-01}  & \num{1.96e-02} $\bullet$\\
 Del & \num{4.28e-01} & \num{2.41e-02} $\bullet$&            \num{5.94e-01}  & \num{2.20e-02} $\bullet$\\
 Ins & \num{5.59e-01} & \num{3.01e-02} $\bullet$&            \num{5.98e-01}  & \num{2.68e-02} $\bullet$\\
 \hline
% Random effects & Variance  & Std. Dev.          & Variance  & Std. Dev.         \\ \hline
% UttID (Intercept)             & \num{0.00573}  & \num{0.0757}   &            \num{0.002584}  & \num{0.05083}  \\
% WER             & \num{0.232861} & \num{0.48256}  &            \num{0.226028}  & \num{0.47542}  \\
% SysID (Intercept)             & 0        & 0        &            0         & 0        \\
% Residual             & \num{0.005293} & \num{0.07275}  &            \num{0.003808}  & \num{0.06171} 
\end{tabular}%
}
\caption[Mixed-effects summary of ASR Word Error Rate as an explanation for MT errors, using Neural MT versus Phrase-based MT.]{Mixed-effects summary, comparing Neural MT (\textsc{neural}) compared to Phrase-based MT (\textsc{mmt}) for SLT. 
Top: WER score as a single predictor of translation $\Delta$TER.
Bottom: Decomposing WER into the basic alignment error operations.
% The model is constructed with observations from all ASR systems in IWSLT 2013's ASR Track.
% Fixed effects coefficients ($\beta$) and standard errors are reported.
% %Statistically significant coefficients ($\beta$) and standard error are reported for fixed effects.
% Random intercepts account for variances by utterance (\textit{UttID}) with a random slope associated with the WER score, and by ASR system (\textit{SysID}).
Statistical significance at $p < 10^{-4}$ is marked with $\bullet$.\vspace{-0.5cm}}
\label{tbl:nmtAnalysis-mixed}
\end{table}

% \section{ASR Error Type distribution}
We compare the average ASR error type frequencies in the FBK ASR utterances where \textsc{neural} or \textsc{mmt} yield a better TER score. 
We introduce the ``phonetic substitution span'' error type from \cite{Ruiz:15:ASRU} to cover multi-word substitution errors (e.g. ``anatomy'' $\Rightarrow$ ``and that to me'').
Focusing on utterances between 10 and 20 words, we observe in Table~\ref{tbl:stat-filter} that the cases where \textsc{neural} scores highest consist of utterances with fewer deletion errors (0.22 versus 0.32). Although further investigation is needed to understand the interplay between substitution and deletion ASR errors in NMT, it is interesting to note that \textsc{mmt} seems to be more adept to handle error-prone ASR outputs, given the higher average WER (19.4\% vs 17.7\%).
% Since the frequencies of insertions and substitutions are roughly the same

\begin{table}[bt]
\footnotesize
\centering
\begin{tabular}{l|r|r|r}
\hline\hline
	&		\textsc{neural}				&		\textsc{mmt}				&		Tie				\\
\hline
Length$_\textit{gold}$	&	$	\num{14.75}	\pm	\num{0.271006469}	$	&	$	\num{14.78481}	\pm	\num{0.367306772}	$	&	$	\num{14.921569}	\pm	\num{0.435523653}	$	\\
WER	&	$	\num{17.7354}	\pm	\num{1.688657697}	$	&	$	\num{19.3856}	\pm	\num{2.535430588}	$	&	$	\num{17.2822}	\pm	\num{2.76947395}	$	\\
\hline
Sub	&	$	\num{1.189655}	\pm	\num{0.139561188}	$	&	$	\num{1.151899}	\pm	\num{0.172884945}	$	&	$	\num{0.19533193
}	\pm	\num{1.117647}	$	\\
Del	&	$	\num{0.215517}	\pm	\num{0.045495822}	$	&	$	\num{0.316456}	\pm	\num{0.06870698}	$	&	$	\num{0.333333}	\pm	\num{0.124196722}	$	\\
Ins	&	$	\num{0.224138}	\pm	\num{0.05207278}	$	&	$	\num{0.265823}	\pm	\num{0.077925163}	$	&	$	\num{0.352941}	\pm	\num{0.121505384}	$	\\
Sub-span	&	$	\num{1.0}	\pm	\num{0.149962541}	$	&	$	\num{0.924051}	\pm	\num{0.185284763}	$	&	$	\num{0.745098}	\pm	\num{0.205499224}	$	\\
\hline
TER & \num{41.9975} vs \num{57.6568965517} & \num{47.7148101266}  vs \num{64.0240506329} & $\num{49.692745098}$ \\
\hline
\end{tabular}
\caption{Average ASR error counts for utterances translated best with \textsc{neural}, \textsc{mmt}, or a tie. 
Translation TER is compared between the best MT system and the inferior MT system.
Computed on utterances with reference (gold) length between 10-20 words.
\vspace{-0.5cm}}
\label{tbl:stat-filter}
\end{table}

\section{Conclusion}
We have introduced a preliminary analysis of the impact of ASR errors on SLT for models trained by neural machine translation systems. In particular, we identify the following as areas to focus on in new research in evaluating NMT for spoken language translation scenarios: (1) contextual patterns not observed during training -- SMT systems usually can back off to shorter sized entries in their translation table; NMT behavior can be erratic. (2) localized and minor ASR errors can cause long distance errors in translation. (3) NMT duplicates content words when minor ASR errors cause the modification of function words.
% We will expand this analysis further to include mixed-effects models that account for errors that can be tolerated with subword units.
Most of the observable errors above are caused by minor substitution errors caused by noisy ASR.
We will expand this analysis further by evaluating NMT architectures that model coverage as well as the representation of inputs with subword units.

\bibliographystyle{IEEEtran}

\bibliography{main.bbl}

% \begin{thebibliography}{9}
% \bibitem[1]{Davis80-COP}
%   S.\ B.\ Davis and P.\ Mermelstein,
%   ``Comparison of parametric representation for monosyllabic word recognition in continuously spoken sentences,''
%   \textit{IEEE Transactions on Acoustics, Speech and Signal Processing}, vol.~28, no.~4, pp.~357--366, 1980.
% \bibitem[2]{Rabiner89-ATO}
%   L.\ R.\ Rabiner,
%   ``A tutorial on hidden Markov models and selected applications in speech recognition,''
%   \textit{Proceedings of the IEEE}, vol.~77, no.~2, pp.~257-286, 1989.
% \bibitem[3]{Hastie09-TEO}
%   T.\ Hastie, R.\ Tibshirani, and J.\ Friedman,
%   \textit{The Elements of Statistical Learning -- Data Mining, Inference, and Prediction}.
%   New York: Springer, 2009.
% \bibitem[4]{YourName17-XXX}
%   F.\ Lastname1, F.\ Lastname2, and F.\ Lastname3,
%   ``Title of your INTERSPEECH 2017 publication,''
%   in \textit{Interspeech 2017 -- 18\textsuperscript{th} Annual Conference of the International Speech Communication Association, August 20?24, Stockholm, Sweden, Proceedings, Proceedings}, 2017, pp.~100--104.
% \end{thebibliography}

\end{document}